%% file: neurips_2025.tex
\documentclass{article}

 \usepackage[dblblindworkshop, final]{neurips_2025}

\usepackage[utf8]{inputenc} 
\usepackage[T1]{fontenc}    
\usepackage{hyperref}       
\usepackage{url}            
\usepackage{booktabs}       
\usepackage{amsfonts}       
\usepackage{nicefrac}       
\usepackage{microtype}      
\usepackage{xcolor}         
\usepackage{amssymb}
\usepackage{graphicx}
\usepackage{algorithm} 
\usepackage{algpseudocode}
\usepackage{tikz}
\usetikzlibrary{
    positioning,   
    arrows.meta,   
    shapes.geometric 
}
\usepackage{booktabs}      
\usepackage{multirow}      
\usepackage{graphicx}      
\usepackage{caption}       
\usepackage{subcaption}    
\usepackage{amsmath}       
\usepackage{pgfplots}      
\pgfplotsset{compat=1.18}
\usepackage{pgf-pie}
\usepackage{wrapfig}

\title{STRIDE: A Systematic Framework for Selecting AI Modalities—Agentic AI, AI Assistants, or LLM Calls}

\author{
  Shubhi Asthana\textsuperscript{1},
  Bing Zhang\textsuperscript{1},
  Chad DeLuca\textsuperscript{1},
  Ruchi Mahindru\textsuperscript{2},
  Hima Patel\textsuperscript{3} \\
  \textsuperscript{1}IBM Research – Almaden, CA, USA \\
  \textsuperscript{2}IBM Research – Yorktown, NY, USA \\
  \textsuperscript{3}IBM Research – India \\
  \texttt{\{sasthan, delucac\}@us.ibm.com},
  \texttt{bing.zhang@ibm.com},
  \texttt{rmahindr@us.ibm.com},
  \texttt{himapatel@in.ibm.com}
}
\begin{document}

\maketitle

\begin{abstract}
The rapid shift from stateless large language models (LLMs) to autonomous, goal-driven agents raises a central question: \textit{When is agentic AI truly necessary?} While agents enable multi-step reasoning, persistent memory, and tool orchestration, deploying them indiscriminately leads to higher cost, complexity, and risk. 

We present \textbf{STRIDE} (Systematic Task Reasoning Intelligence Deployment Evaluator), a framework that provides principled recommendations for selecting between three modalities: (i) direct LLM calls, (ii) guided AI assistants, and (iii) fully autonomous agentic AI. STRIDE integrates structured task decomposition, dynamism attribution, and self-reflection requirement analysis to produce an \textit{Agentic Suitability Score}, ensuring that full agentic autonomy is reserved for tasks with inherent dynamism or evolving context.


Evaluated across \textit{30 real-world tasks} spanning SRE, compliance, and enterprise automation, STRIDE achieved \textit{92\% accuracy} in modality selection, reduced unnecessary agent deployments by \textit{45\%}, and cut resource costs by \textit{37\%}. Expert validation over six months in SRE and compliance domains confirmed its practical utility, with domain specialists agreeing that STRIDE effectively distinguishes between tasks requiring simple LLM calls, guided assistants, or full agentic autonomy. This work reframes agent adoption as a \textit{necessity-driven} design decision, ensuring autonomy is applied only when its benefits justify the costs.


\end{abstract}

\input{introduction} 
\input{related_work}
\input{methodology}

\input{experiments}
\input{conclusion}

\small  

\bibliographystyle{plainnat}
\bibliography{bib}

\end{document}

%% file: introduction.tex
\section{Introduction}
\label{sec:intro}

Recent advances have transformed AI from simple stateless LLM calls to sophisticated autonomous agents, enabling richer reasoning, tool use, and adaptive workflows. While this progression unlocks significant value in domains such as site reliability engineering (SRE), compliance, and automation, it also introduces substantial trade-offs in cost, complexity, and risk. A central design challenge emerges: \textit{when agents are truly necessary}, and when are simpler alternatives sufficient?


We distinguish three modalities: (i) \textbf{LLM calls}, providing single-turn inference without memory or tools, which is ideal for straightforward query-response scenarios; (ii) \textbf{AI assistants}, which handle guided multi-step workflows with short-term context and limited tool access that is suitable for structured processes requiring human oversight; and (iii) \textbf{Agentic AI}, which autonomously decomposes tasks, orchestrates tools, and adapts with minimal oversight, which is necessary for complex, dynamic environments requiring independent decision-making. Table~\ref{tab:ai_modalities} contrasts these modalities. 

Current practice often overuses agentic AI, deploying autonomous systems even when simpler modalities would suffice. This tendency leads to unnecessary cost, complexity, and risk, particularly in enterprise contexts where reliability and governance are critical. \textit{A principled framework for deciding when agents are truly necessary has been missing,} leaving design-time choices largely intuition-driven rather than evidence-based. While agentic AI unlocks transformative value in domains like SRE, compliance verification, and complex automation, deploying it indiscriminately carries risks:
\begin{itemize}
    \item \textbf{Overengineering:} using agents for simple queries wastes compute and developer effort.
    \item \textbf{Security \& compliance risks:} uncontrolled tool use and API calls may leak sensitive data.
    \item \textbf{System instability:} recursive loops and unbounded workflows degrade reliability.
\end{itemize}


\begin{table}[t]
    \centering
    \caption{Comparison of AI Modalities}
    \label{tab:ai_modalities}
    \begin{tabular}{lcccc}
        \toprule
        \textbf{Attribute} & \textbf{LLM Call} & \textbf{AI Assistant} & \textbf{Agentic AI} \\
        \midrule
        Reasoning Depth & Shallow & Medium & Deep \\
        Tool Needs & Single & Single/Multiple & Multiple \\
        State Needs & None & Ephemeral & Persistent \\
        Risk Profile & Low & Medium & High \\
        Use Case Example & Exchange rate lookup & Summarize meeting notes & Plan 5-day travel itinerary \\
        \bottomrule
    \end{tabular}
\end{table}

We propose \textbf{STRIDE}, a novel framework for \textit{necessity assessment at design time}: systematically deciding whether a given task should be solved with an LLM call, an AI assistant, or agentic AI. STRIDE analyzes task descriptions across four integrated analytical dimensions:

\begin{itemize}
    \item \textbf{Structured Task Decomposition:} Tasks are decomposed into a directed acyclic graph (DAG) of subtasks, systematically breaking down objectives to reveal inherent complexity, interdependencies, and sequential reasoning requirements that distinguish simple queries from multi-step challenges.
    \item \textbf{Dynamic Reasoning and Tool-Interaction Scoring:} STRIDE quantifies reasoning depth together with tool dependencies, external data access, and API requirements, identifying when sophisticated orchestration beyond basic language processing is necessary.
    \item \textbf{Dynamism Attribution Analysis:} Using a \textit{True Dynamism Score (TDS)}, the framework attributes variability to models, tools, or workflow sources, clarifying when persistent memory and adaptive decision-making are required.
    \item \textbf{Self-Reflection Requirement Assessment:} Assesses need for error recovery and meta-cognition, and integrates all factors into an \textit{Agentic Suitability Score (ASS)} that guides the choice of LLM call, assistant, or agent.
\end{itemize}

This unified methodology ensures that AI solution selection is not an ad-hoc judgment call, but a structured, repeatable process that balances capability requirements with efficiency, cost, and risk management. Just as scaling laws have guided model development by quantifying performance as a function of parameters and data, we argue that analogous principles are needed for \textit{environmental and task scaling}. Not every task requires autonomy: simple queries map to LLM calls, structured processes to guided assistants, and only dynamic, evolving workflows demand full agentic AI. STRIDE introduces such a structured scaling perspective for modality selection.

\textbf{Strategic Integration and Impact:} STRIDE acts as a \textit{“shift-left”} decision tool— i.e., it moves critical choices from deployment time to the design phase—embedding modality selection into early workflows. This prevents over-engineering, avoids under-provisioning, and provides defensible criteria for balancing capability, efficiency, computational cost, and risk.

\begin{itemize}
    \item We introduce \textbf{STRIDE}, the first design-time framework for AI modality selection, shifting decisions left in the pipeline.
    \item We define a novel quantitative \textbf{Agentic Suitability Score} with dynamism attribution, balancing autonomy benefits against cost and risk.
    \item We evaluate STRIDE on 30 real-world tasks across SRE \cite{jha2025itbench}, compliance, and enterprise automation, demonstrating reduced agentic over-deployment by \textbf{45\%} while improving expert alignment by \textbf{27\%}.
\end{itemize}

Beyond efficiency, this framing directly supports responsible AI deployment. By preventing over-engineering, STRIDE reduces unnecessary surface area for errors, governance failures, and hidden costs, while ensuring that truly complex tasks receive the level of autonomy they demand. 

%% file: related_work.tex
\section{Related Work}
\label{sec:related}
Recent advances have expanded AI from simple LLM calls to guided assistants and adaptive agentic systems. While assistants follow structured workflows, agents plan and make inference-time decisions in dynamic environments. This shift has driven research into task complexity, reasoning depth, and self-reflection, but few works address the design-time question of \textit{when agents are truly needed}. Related work such as AgentBoard \cite{chang2024agentboard} benchmarks multi-turn agent evaluation via task decomposition and error taxonomy, aligning with STRIDE’s scoring. COPPER \cite{bo2024reflective} introduces self-reflection via counterfactual rewards in multi-agent settings, reinforcing the role of reflection analysis in STRIDE. While frameworks address components of intelligent execution \cite{ye2024efficient,kapoor2024ai}, few offer a systematic methodology for selecting the appropriate AI modality at design time. 

\paragraph{Benchmarks for agent performance.} A growing body of benchmarks evaluates how well agents perform specific tasks. AgentBench \cite{xu2025towards}, ITBench \cite{jha2025itbench}, and ToolBench \cite{qin2025meta} stress-test multi-tool reasoning and environment interaction. SWE-Bench \cite{jimenez2023swe} focuses on software engineering workflows, while Gorilla \cite{patil2024gorilla} evaluates large-scale tool invocation. HuggingGPT \cite{shen2023hugginggpt} and ReAct \cite{yao2023react} integrate tool usage and reasoning traces to improve robustness. These works emphasize \textit{performance measurement after deployment}. By contrast, STRIDE addresses the orthogonal but complementary question of \textit{necessity at design time}: before deploying agents, can we predict whether a task truly requires them?

\paragraph{Task complexity and modality selection.} Prior studies classify tasks for LLMs, assistants, or agents: agents excel at workflow decomposition but risk loops \cite{ibm_agents_vs_assistants_2025}; small LMs suit repetitive subtasks \cite{belcak_small_lms_2025,nvidia_slms_2025}; and governance risks remain a concern \cite{mckinsey_agentic_ai_2025}. STRIDE formalizes these intuitions into a scoring framework that balances reasoning depth, tool needs, and state requirements.

\paragraph{Task decomposition, Self-reflection and adaptive reasoning.} Decomposition is central: graph-based metrics support evaluation \cite{gabriel_advancing_agentic_2024}; TDAG automates subtasks \cite{crispino_tdag_2025}; and tool-calling studies quantify volatility from nested or parallel use \cite{masterman_tool_calling_2024}, factors we incorporate in the True Dynamism Score. Reflection has been explored in ARTIST \cite{plaat_artist_2025} and MTPO \cite{wu_large_reasoning_models_2025}. We instead treat reflection as a necessity criterion rather than a performance add-on.

\paragraph{Industry and patents.} Frameworks such as LlamaIndex, Google ADK, and CrewAI \cite{agentic_ai_frameworks_2025} enable modular workflows, while patents from Anthropic and OpenAI \cite{patent_ai_workflows_2024,afp_openai_operator_2025} describe autonomous travel and compliance. STRIDE differs by focusing on \textit{design-time necessity assessment}, embedding explainability and risk-awareness into early choices.

While prior work evaluates agent capabilities post-deployment, no framework automates modality selection \textit{at design time}. STRIDE fills this gap with task complexity scoring, variability attribution, drift monitoring, and persona-specific recommendations, uniquely addressing the question of \textit{whether agents are needed at all} and transforming solution selection into a structured, evidence-based discipline.

%% file: methodology.tex
\section{Methodology }
\label{sec:methodology}

\begin{figure}[t]
    \centering
    \includegraphics[width=\linewidth]{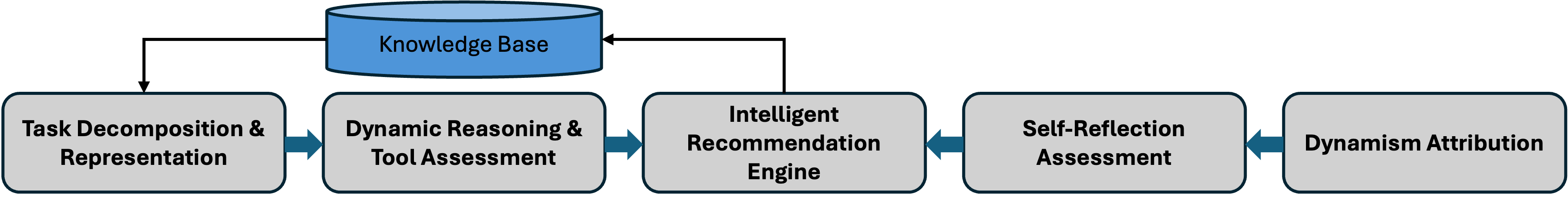}
    \caption{
        Overview of STRIDE, a five-stage framework for determining the necessity of Agentic AI, AI assistants, or LLM calls.
        Stage 1: Task decomposition into subtasks with dependency graph construction.
        Stage 2: Dynamic reasoning and tool-interaction scoring.
        Stage 3: Dynamism attribution (model/tool/workflow).
        Stage 4: Self-reflection requirement analysis.
        Stage 5: Aggregated suitability inference with persona-aware recommendations.
    }
    \label{fig:flow}
\end{figure}

In this section, we present our end-to-end framework, STRIDE (Systematic Task Reasoning Intelligence Deployment Evaluator), for assessing whether a task requires the deployment of \textit{agentic AI}, an \textit{AI assistant}, or a \textit{stateless LLM call}. STRIDE systematically evaluates \textit{\textbf{task complexity}}, \textit{\textbf{reasoning depth}}, \textit{\textbf{tool dependencies}}, \textit{\textbf{dynamism of task}}, and \textit{\textbf{self-reflection}} requirements to provide a quantitative recommendation. Figure~\ref{fig:flow} illustrates the workflow

\subsection{System Overview}
STRIDE analyzes task descriptions, inputs/outputs, and tool dependencies to recommend the appropriate AI modality. This process comprises producing an \textit{Agentic Suitability Score (ASS)} for each subtask. This score is then aggregated to guide the final modality recommendation:
\begin{itemize}
    \item \textbf{Task Decomposition:} Breaks tasks into a DAG of subtasks to expose dependencies.
    \item \textbf{Reasoning \& Tool Scoring:} Quantifies reasoning depth, tool reliance, and API orchestration requirements.
    \item \textbf{Dynamism Analysis:} Attributes variability across model, tool, and workflow sources using a True Dynamism Score (TDS) to determine whether adaptive agentic reasoning is needed.
    \item \textbf{Self-Reflection Assessment:} Detects when iterative correction is required and integrates all factors into an \textit{Agentic Suitability Score (ASS)} to give final recommendation.
\end{itemize}

\subsection{Task Decomposition \& Representation}
In this stage, STRIDE transforms free-form task descriptions into structured, actionable subtasks using a fine-tuned LLM with specialized prompting. The system identifies key action verbs (like "search," "validate," "analyze") and target nouns (such as "flights," "budget," "data") to create meaningful work units.
To illustrate with a practical example, if the initial task is "Plan a 5-day travel itinerary", the Task Decomposition phase would generate subtasks like "Search Flights", "Find Hotels", "Budget Planning", and "Activity Research".

The system automatically discovers relationships between subtasks through 1) \textit{Temporal Analysis:} Recognizing sequence requirements ("search flights before booking hotels"), 2) \textit{Data Flow Tracking:} Identifying when one subtask's output feeds into another ("Search Flights" results inform "Budget Alerts"), and 3) \textit{Semantic Role Labeling:} Mapping precise input/output relationships.

STRIDE creates a directed acyclic graph (DAG) where each subtask node contains, 1) \textit{Historical Patterns:} "Search Flights" appears as the starting point in 85\% of travel planning tasks, 2) \textit{Tool Recommendations:} Proven integrations for similar subtasks, and 3) \textit{Performance Insights:} Success rates and optimization guidance from past executions. By converting ambiguous requests into precise, interconnected subtasks, STRIDE establishes the foundation for intelligent automation decisions. This structured approach ensures no critical dependencies are missed while enabling parallel execution where possible.
Let $T = \{s_1, s_2, \ldots, s_n\}$ represent the extracted subtasks, organized in graph $G = (T, E)$ where edges $E$ capture both ordering constraints and data dependencies between tasks.

\subsection{Dynamic Reasoning \& Tool Assessment}
For each subtask $s_i$, STRIDE computes an \textit{Agentic Suitability Score} (ASS) that objectively measures whether the subtask benefits from autonomous agent capabilities:
\begin{equation}
    \text{ASS}(s_i) = w_r \cdot R(s) + w_t \cdot T(s) + w_s \cdot S(s) + w_\rho \cdot \rho(s),
\end{equation}
where:
\begin{itemize}
    \item $R(s)$ = Reasoning depth (0 = \textit{Shallow}; simple lookup or direct response, 1 = \textit{Medium}; requires comparison or basic inference, 2 = \textit{Deep}; multi-step analysis or complex decision-making),
    \item $T(s)$ = tool need (0 = \textit{None}; no external tools required, 1 = \textit{Single}; single tool integration, 2 = \textit{Multiple}; multiple tool orchestration needed),
    \item $S(s)$ = state/memory requirement (0 = \textit{None}; stateless operation, 1 = \textit{Ephemeral}; single session, 2 = \textit{Persistent};),
    \item $\rho(s)$ = Risk Score (compliance violations, computational Overhead, infinite loop potential).
\end{itemize}
The weighting system $(w_r, w_t, w_s, w_\rho)$ adapts to different task domains: 
\textit{Reasoning-Heavy Tasks:} \(w_r\)) prioritizes complex multi-step tasks (e.g., \(w_r = 0.4\) for itinerary planning)
\textit{Tool-Intensive Workflows:} (\(w_t\)) emphasizing tasks requiring multiple tools (e.g., \(w_t = 0.3\) for API-heavy workflows)
\textit{Context-Dependent Operations:} (\(w_s\)) accounting for persistent context needs (e.g., \(w_s = 0.2\) for multi-turn interactions)
\textit{Risk-Sensitive Applications:} (\(w_\rho\)), penalizing high-risk operations (e.g., \(w_\rho = 0.1\) for compliance tasks)

STRIDE continuously refines these weights through \textit{grid search optimization} on labeled historical task data, then refines via \textit{reinforcement learning} from deployment outcomes and \textit{expert feedback integration} for domain-specific calibration. This scoring mechanism prevents over-engineering simple tasks with complex agentic AI solutions, while ensuring that sophisticated problems receive appropriate autonomous capabilities. The result is precise resource allocation and optimal performance across diverse task types.


\subsection{Dynamism Attribution}
Variability alone does not justify implementing AI agents. For instance, a task like *"Generate a random greeting message"* may produce different outputs each time due to model stochasticity (model-induced variability), but it can be handled effectively by a stateless LLM with temperature adjustments—no agentic autonomy is required.
 STRIDE distinguishes:
\begin{itemize}
    \item \emph{Model-induced variability}, 
    stems from AI model limitations, including prompt ambiguity (unclear prompts causing inconsistent outputs) and stochastic randomness (probabilistic models producing different results from identical inputs). This variability typically resolves through improved prompt engineering, temperature controls, or deterministic sampling rather than requiring agentic capabilities.
    \item \emph{Tool-induced variability}, 
    arises from external dependencies, including API volatility (changing response formats, rate limits, downtime) and dynamic tool responses (varying data based on real-time conditions). These challenges typically require robust error handling, retry mechanisms, and adaptive response parsing rather than autonomous agent decision-making.
    \item \emph{Workflow-induced variability}, 
    involves systemic execution complexity, including conditional branching (different inputs triggering varied decision trees) and environmental changes (system load, user context, data availability altering optimal paths). This category most strongly indicates agentic solution needs, as it requires dynamic decision-making and adaptive planning that benefit from autonomous reasoning capabilities.
\end{itemize}

By distinguishing sources of variability, STRIDE avoids over-engineering and activates agentic AI only when autonomous reasoning materially improves task outcomes.

The \emph{True Dynamism Score (TDS)} isolates workflow-driven variability:
\begin{equation}
    \text{TDS}(s_i) = \alpha \cdot W(s) + \beta \cdot V(s) - \gamma \cdot M(s),
\end{equation}
where $W(s)$ is workflow variability, $V(s)$ tool volatility, and $M(s)$ model instability. A high TDS implies that autonomy and adaptivity are required.

\subsection{Self-Reflection Assessment}
Self-reflection is required when subtasks involve mid-execution decision points or validation of nondeterministic tools. 

\textbf{Mid-execution decision points} occur when workflows cannot be fully predetermined and require dynamic evaluation during execution. AI Agents implement procedural mechanisms to incorporate tool responses mid-process, while Agentic AI introduces recursive task reallocation and cross-agent messaging for emergent decision-making \cite{sapkota2025ai}. These situations arise when initial conditions change unexpectedly, multi-step processes reveal information influencing subsequent actions, or quality checkpoints require evaluating whether intermediate outputs meet success criteria. The Reflexion framework demonstrates how agents reflect on task feedback and maintain reflective text in episodic memory to improve subsequent decision-making \cite{shinn2023reflexion}, with studies showing significant problem-solving performance improvements (p < 0.001) \cite{renze2024self}.

\textbf{Validation of nondeterministic tools} becomes critical when working with external systems producing variable outputs. LLM-powered systems present challenges where outputs are unpredictable, requiring custom validation frameworks. This includes API responses with different data structures, LLM-generated content requiring accuracy evaluation, and web scraping tools exhibiting behavior changes due to evolving website structures. Neural network instability can lead to disparate results, requiring rigorous validation through adversarial robustness testing.

Without self-reflection, agents risk propagating errors, making incorrect assumptions about tool outputs, or failing to adapt when strategies prove insufficient. Self-reflection enables task coherence and reliability in dynamic environments. STRIDE encodes this as a decision rule:
\[
\text{SR}(s) = \mathbf{1}\!\left(\text{TDS}(s) \geq \theta \land \left(C(s) \lor N(s) \lor V(s)\right)\right),
\]
where $C(s)$ = conditional branches, $N(s)$ = nondeterministic tools, $V(s)$ = mid-execution validation, and $\theta$ = dynamism threshold. If $\text{SR}(s) = 1$, reflection hooks (e.g., error recovery, re-planning, ReAct) are triggered.

\begin{wrapfigure}{r}{0.5\textwidth} 
    \centering
    \includegraphics[width=0.48\textwidth]{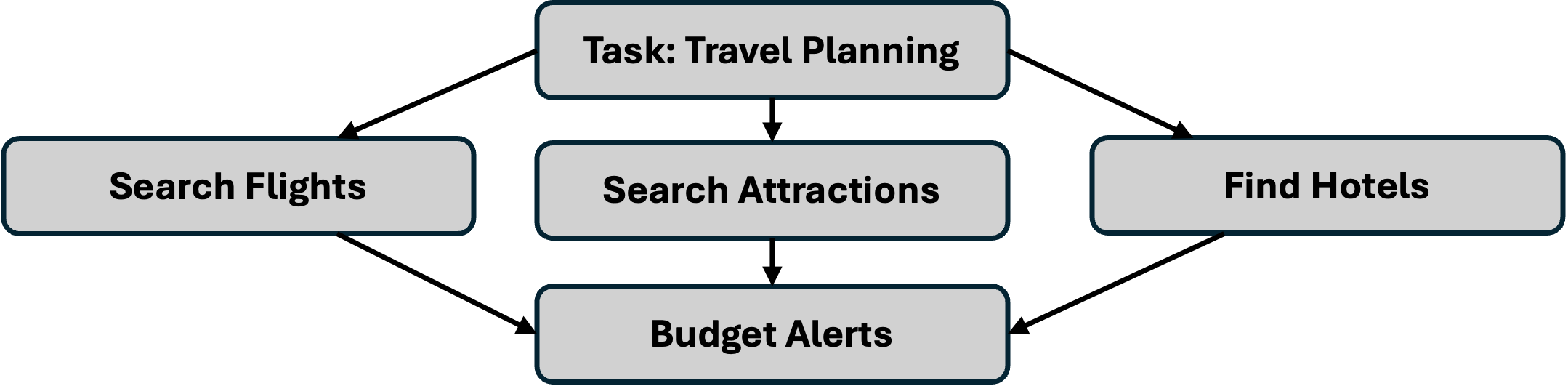} 
    \caption{
        Toy decomposition DAG for ``Plan 5-day travel itinerary.''
        Each subtask is scored separately and orchestrated by STRIDE.
    }
    \label{fig:toy_dag}
\end{wrapfigure}

\subsection{Intelligent Recommendation Engine}
Finally, STRIDE aggregates features from subtasks into a task profile $\mathbf{x}_T$ and queries a knowledge base $\mathcal{K}$ of historical patterns. A classifier $f$ produces the final modality:
\begin{equation}
    \hat{y} = \arg\max_{m \in \{\text{LLM}, \text{Assistant}, \text{Agent}\}} f(\mathbf{x}_T; \mathcal{K}),
\end{equation}
with justification tailored to the user’s persona (e.g., developers receive tool configurations, managers receive architectural summaries).

\begin{algorithm}[t]
\caption{STRIDE Scoring \& Modality Inference}
\label{alg:stride}
\begin{algorithmic}[1]
\Require Task description \( \tau \), knowledge base \( \mathcal{K} \), thresholds \( \theta, \dots \)
\Ensure Modal suggestion \( \hat{y} \in \{\texttt{LLM\_CALL}, \texttt{AI\_ASSISTANT}, \texttt{AGENTIC\_AI}\} \)
\State Decompose \( \tau \) into subtasks \( T = \{s_1, \dots, s_n\} \) and build DAG \( G \)
\For{each subtask \( s \in T \)}
  \State Compute \( R(s), T(s), S(s), \rho(s) \) and derive \( \text{ASS}(s) \)
  \State Compute \( W(s), V(s), M(s) \) and derive \( \text{TDS}(s) \)
  \State Evaluate \( C(s), N(s), V(s) \) to derive \( \text{SR}(s) \)
\EndFor
\State Aggregate features into task profile \( \mathbf{x}_T \)
\State Return \( \hat{y} = \arg\max_{m} f(\mathbf{x}_T; \mathcal{K}) \)
\end{algorithmic}
\end{algorithm}

Figure \ref{fig:toy_dag} illustrates a toy DAG for a travel-planning task, showing how STRIDE decomposes tasks into subtasks for scoring and routing. To clarify the STRIDE workflow, Algorithm \ref{alg:stride} outlines the end-to-end scoring and modality inference process, from task decomposition to final recommendation. This structured scoring-to-classification pipeline ensures that agentic AI is deployed only when justified by objective complexity, resource trade-offs, and dynamism.

%% file: experiments.tex
\section{Experiments \& Results}
\label{sec:experiments}

We evaluated STRIDE across 30 real-world tasks spanning SRE, enterprise automation, legal compliance, and customer support. The objective was to test whether STRIDE reliably distinguishes between LLM calls, assistants, and agents, minimizing over-engineering while ensuring accurate, cost-efficient design-time decisions. While modest in size, our task set emphasizes \textit{depth over breadth}, demonstrating STRIDE’s value in real-world settings. Across all 30 tasks, STRIDE achieved \textbf{92\% accuracy}, reduced unnecessary agent deployments by \textbf{45\%}, and delivered \textbf{37\% lower compute/API usage} compared to always deploying agents. \textbf{\textit{These results demonstrate that principled design-time selection yields tangible efficiency gains compared to intuition-driven deployment.}} We compared STRIDE against two baselines. The \textbf{Naive Agent} baseline always deployed agentic AI regardless of task complexity, providing an upper bound on cost but no efficiency. The \textbf{Heuristic Threshold} baseline deployed agents only when reasoning depth $\geq 2$ and tool requirements $\geq 2$, but often failed on borderline cases where task dynamism or reflection was the deciding factor. STRIDE consistently outperformed both approaches.

\begin{table}[t]
    \centering
    \caption{Quantitative results of STRIDE compared to baselines across 30 tasks.}
    \label{tab:quant_results}
    \begin{tabular}{lccc}
        \toprule
        \textbf{Method} & \textbf{Accuracy (\%)} & \textbf{Over-engg Reduction (\%)} & \textbf{Resource Savings (\%)} \\
        \midrule
        Naive Agent & 33.3 & 0 & 0 \\
        Heuristic Threshold & 68.0 & 27.5 & 18.2 \\
        STRIDE (ours) & \textbf{92.0} & \textbf{45.3} & \textbf{37.1} \\
        \bottomrule
    \end{tabular}
\end{table}

\subsection{Illustrative Use Cases}
To ground these aggregate results, we highlight representative tasks where STRIDE discriminates between simple lookups, medium-complexity assistance, and fully autonomous agent workflows. These cases illustrate how STRIDE’s scoring pipeline translates into practical deployment recommendations.

\paragraph{LLM Call Example: Currency Lookup.}  
\textit{``What is the exchange rate between USD and EUR today?''}  
This task requires shallow reasoning (0-hop), a single API call, and no state persistence. STRIDE assigned a low True Dynamism Score (0.10) and recommended \texttt{LLM\_CALL}. This minimized cost and latency, avoiding unnecessary orchestration overhead while retaining accuracy.

\paragraph{AI Assistant Example: Meeting Summarization.}  
\textit{``Summarize today’s team meeting notes and suggest action items.''}  
This task requires medium reasoning depth (1-hop), a summarization tool, and ephemeral state. STRIDE produced a TDS of 0.35 and recommended \texttt{AI\_ASSISTANT}, reflecting that autonomy is unnecessary but structured guidance improves usability. Deploying a fully autonomous agentic AI for this task would have added unnecessary computation and orchestration overhead without improving the outcome, since an AI assistant sufficed.

\paragraph{Agentic AI Example: Travel Planning.}  
\textit{``Plan a 5-day travel itinerary with hotels, attractions, and budget alerts.''}  
This task demands multi-hop reasoning, persistent state, and multiple API integrations (flights, hotels, maps). STRIDE assigned a TDS of 0.78 and correctly recommended \texttt{AGENTIC\_AI}. Experts validated that dynamic replanning is essential in such workflows due to evolving constraints and interdependencies.

\paragraph{SRE Example: Kubernetes Incident Analysis.}  
\textit{``Analyze Kubernetes change events and correlate them with active alerts to identify the root cause of an ongoing incident.''}  
This high-stakes task requires deep reasoning, multiple tool integrations (Kubernetes API, alerting system, causal analysis), and persistent state tracking. STRIDE scored a TDS of 0.85 and recommended \texttt{AGENTIC\_AI}. Domain experts confirmed that incident resolution often requires iterative exploration and adaptive strategies that static assistants cannot provide.

\paragraph{Compliance Verification Example.}  
\textit{``Evaluate a set of documents for legal compliance, flagging any non-compliant sections and suggesting corrections.''}  
This task involves deep reasoning, persistent state, and multiple specialized tools (legal database, document parser, compliance checker). STRIDE assigned a TDS of 0.80 and recommended \texttt{AGENTIC\_AI}, reflecting the high compliance risks and iterative refinements required. Experts noted that assistants often fail to capture edge cases in regulatory contexts.

\begin{table}[t]
    \centering
    \caption{Representative task evaluations. RD = Reasoning Depth, TN = Tool Needs, SN = State Needs, TDS = True Dynamism Score.}
    \label{tab:results_examples}
    \begin{tabular}{lcccccc}
        \toprule
        \textbf{Task} & \textbf{RD} & \textbf{TN} & \textbf{SN} & \textbf{TDS} & \textbf{Risk} & \textbf{Recommendation} \\
        \midrule
        Currency lookup & 0 & 1 & 0 & 0.10 & Low & LLM\_CALL \\
        Meeting summarization & 1 & 1 & 1 & 0.35 & Medium & AI\_ASSISTANT \\
        Travel itinerary planning & 2 & 2 & 2 & 0.78 & High & AGENTIC\_AI \\
        Kubernetes incident analysis & 2 & 2 & 2 & 0.85 & High & AGENTIC\_AI \\
        Legal compliance verification & 2 & 2 & 2 & 0.80 & High & AGENTIC\_AI \\
        \bottomrule
    \end{tabular}
\end{table}

\subsection{Why STRIDE Works: Ablation Study}
To understand why STRIDE performs well, we conducted ablation experiments by removing core components. As Table~\ref{tab:ablation_integrated} shows, each element contributes significantly. Removing task decomposition reduced accuracy by 9\%, showing that subtask structure is essential for modeling dependencies. Without the True Dynamism Score, accuracy fell by 12\%, as STRIDE struggled to distinguish borderline tasks like meeting summarization versus compliance verification. The largest drop came from removing self-reflection, which reduced accuracy to 76\%, underscoring its role in handling mid-execution corrections and adaptive reasoning. 

Human-in-the-loop validation also played a role: omitting expert feedback reduced alignment with domain judgments, demonstrating the value of incorporating expert calibration into design-time recommendations.

\begin{table}[t]
\centering
\caption{Ablation study of STRIDE components. Accuracy, over-engineering reduction, and resource savings are reported.}
\label{tab:ablation_integrated}
\begin{tabular}{lccc}
\toprule
\textbf{Configuration} & \textbf{Accuracy (\%)} & \textbf{Over-engg Reduction (\%)} & \textbf{Resource Savings (\%)} \\
\midrule
Full STRIDE & \textbf{92.0} & \textbf{45.3} & \textbf{37.1} \\
w/o Task Decomposition & 83.0 & 35.2 & 28.0 \\
w/o True Dynamism Score & 80.0 & 33.0 & 26.5 \\
w/o TDS Weighting & 81.3 & 32.0 & 25.4 \\
w/o Self-Reflection & 76.0 & 29.5 & 22.8 \\
w/o Human-in-the-loop & 85.7 & 37.1 & 28.6 \\
\bottomrule
\end{tabular}
\end{table}

\subsection{Robustness and Human Validation}
Beyond aggregate numbers, we tested robustness across domains. STRIDE achieved 95\% accuracy in SRE, 91\% in compliance, 89\% in automation, and 93\% in customer support (Figure~\ref{fig:domain_accuracy}). This consistency suggests that STRIDE generalizes well across heterogeneous real-world tasks without overfitting to any specific domain. Errors primarily arose in borderline scenarios, such as multi-document summarization, where dynamism was underestimated. Notably, STRIDE sometimes recommended assistants when experts preferred agents, but never the reverse—avoiding costly over-engineering mistakes.

Expert validation further confirmed STRIDE’s recommendations. In 78\% of cases, experts fully agreed, 15\% showed partial agreement (e.g., suggesting an assistant instead of an agent for borderline tasks), and only 7\% disagreed (Figure~\ref{fig:expert_agreement}). This resulted in a \textbf{27\%} improvement in expert alignment compared to the Heuristic Threshold baseline. Feedback from engineers and compliance officers improved STRIDE through better task decomposition, adjusted TDS weights, and persona-aware outputs tailored to developers and managers (Table~\ref{tab:feedback_improvements}). Our robustness validation was not a one-off annotation exercise, but the result of extended collaboration with subject matter experts. 
For the SRE domain, three Kubernetes incident response experts engaged with STRIDE iteratively over a six-month period (March–August 2025), providing feedback on decomposition, reflection, and dynamism scoring. 
In the compliance domain, two legal verification experts participated in a shorter but focused engagement of 1–2 months (May-June 2025), helping calibrate task scoring against regulatory criteria. This sustained, multi-month collaboration ensured that STRIDE’s assessments aligned with the nuanced realities of enterprise practice.

\begin{figure}[t]
    \centering
    \begin{minipage}[t]{0.45\linewidth}
        \centering
        \begin{tikzpicture}
            \begin{axis}[
                ybar,
                bar width=12pt,
                ylabel={Accuracy (\%)},
                symbolic x coords={SRE, Compliance, Automation, Support},
                xtick=data,
                ymin=0, ymax=100,
                nodes near coords,
                nodes near coords align={vertical},
                width=\linewidth,
                height=5.5cm,
                enlarge x limits=0.15,
                ymajorgrids=true,
                grid style=dashed,
                x tick label style={rotate=30, anchor=east, font=\small},
            ]
            \addplot coordinates {(SRE,95) (Compliance,91) (Automation,89) (Support,93)};
            \end{axis}
        \end{tikzpicture}
        \caption{Domain-wise accuracy of STRIDE across 30 tasks.}
        \label{fig:domain_accuracy}
    \end{minipage}
    \hfill
    \begin{minipage}[t]{0.45\linewidth}
        \centering
        \begin{tikzpicture}
            \pie[
                text=legend,
                radius=2,
                color={blue!50, red!30, green!50}
            ]{
                78/Full Agreement,
                15/Partial Agreement,
                7/Disagreement
            }
        \end{tikzpicture}
        \caption{Expert agreement with STRIDE recommendations.}
        \label{fig:expert_agreement}
    \end{minipage}
\end{figure}

\begin{table}[t]
    \centering
    \caption{Summary of Human-in-the-Loop Feedback and System Improvements.}
    \label{tab:feedback_improvements}
    \begin{tabular}{lp{10cm}}
        \toprule
        \textbf{Feedback Area} & \textbf{Improvements Made} \\
        \midrule
        Task Decomposition & Enhanced LLM-driven decomposition to better capture subtask dependencies. \\
        Dynamism Analysis & Adjusted weights in the True Dynamism Score to better separate model-, tool-, and workflow-induced variability. \\
        Knowledge Base & Expanded task patterns and historical performance metrics for SRE and compliance tasks. \\
        \bottomrule
    \end{tabular}
\end{table}

\subsection{Discussion and Limitations}
STRIDE reduces the costs, risks, and misaligned expectations of unnecessary agents. By shifting selection to design time, it prevents over-engineering, ensures autonomy only where required, and reframes adoption from intuition-driven to structured decision process that directly translates into lower compute/API expenditure and reduced operational costs. At the same time, we acknowledge limitations. STRIDE’s scoring functions are heuristic by design, striking a balance between interpretability and generality. 

Finally, STRIDE complements existing benchmarks, such as AgentBench, SWE-Bench, and ToolBench. While those benchmarks evaluate \textit{how well} agents perform after deployment, STRIDE focuses on \textit{whether agents are needed at all} before deployment. 
This creates opportunities for integration: STRIDE could serve as a design-time filter that guides which tasks should be benchmarked with agents, or as a planning tool embedded into enterprise AI workflows. 
Together, these directions position STRIDE as both a practical engineering aid and a guardrail for responsible AI deployment.

%% file: conclusion.tex
\section{Conclusion}
\label{sec:conclusion}
We introduced STRIDE (Systematic Task Reasoning Intelligence Deployment Evaluator), a framework for systematically determining when tasks require agentic AI, AI assistants, or simple LLM calls. STRIDE integrates five analytical dimensions — structured task decomposition, dynamic reasoning and tool-interaction scoring, dynamism attribution analysis, self-reflection requirement assessment, and agentic suitability inference. In evaluating 30 real-world enterprise tasks, STRIDE reduced unnecessary agent deployments by 45\%, improved expert alignment by 27\% and cut resource costs by 37\%, directly mitigating over-engineering risks and containing compute costs.

Looking ahead, we will extend evaluation beyond the 30 tasks to include multimodal tasks (vision/audio), integrate reinforcement learning for weight tuning, and validate STRIDE at enterprise scale. These extensions will further strengthen its role as a practical guardrail for responsible AI deployment.


%% file: bib.bib
@misc{ibm_agents_vs_assistants_2025,
    title = {AI Agents vs. AI Assistants},
    url = {https://www.ibm.com/think/topics/ai-agents-vs-ai-assistants},
    author = {{IBM}},
    year = {2025},
    month = {7},
    urldate = {2025-08-20}
}

@article{belcak_small_lms_2025,
    title = {Small Language Models are the Future of Agentic AI},
    author = {Belcak, Peter and others},
    journal = {arXiv},
    year = {2025},
    month = {6},
    url = {https://arxiv.org/abs/2506.02153},
    urldate = {2025-08-20}
}

@misc{nvidia_slms_2025,
    title = {NVIDIA Says Small Language Models Are The Future of Agentic AI},
    author = {{Greyling, Cobus}},
    url = {https://cobusgreyling.medium.com/nvidia-says-small-language-models-are-the-future-of-agentic-ai-f1f7289d9565},
    year = {2025},
    month = {6},
    urldate = {2025-08-20}
}

@misc{mckinsey_agentic_ai_2025,
    title = {Seizing the agentic AI advantage},
    author = {{McKinsey \& Company}},
    url = {https://www.mckinsey.com/capabilities/quantumblack/our-insights/seizing-the-agentic-ai-advantage},
    year = {2025},
    month = {6},
    urldate = {2025-08-20}
}

@article{gabriel_advancing_agentic_2024,
    title = {Advancing Agentic Systems: Dynamic Task Decomposition, Tool Integration and Evaluation using Novel Metrics and Dataset},
    author = {Gabriel, Adrian Garret and others},
    journal = {arXiv},
    year = {2024},
    month = {10},
    url = {https://arxiv.org/abs/2410.22457},
    urldate = {2025-08-20}
}

@article{crispino_tdag_2025,
    title = {TDAG: A multi-agent framework based on dynamic Task Decomposition and Agent Generation},
    author = {Crispino, M. and others},
    journal = {ScienceDirect},
    year = {2025},
    month = {1},
    url = {https://www.sciencedirect.com/science/article/abs/pii/S0893608025000796},
    urldate = {2025-08-20}
}

@article{chang2024agentboard,
  title={Agentboard: An analytical evaluation board of multi-turn llm agents},
  author={Chang, Ma and Zhang, Junlei and Zhu, Zhihao and Yang, Cheng and Yang, Yujiu and Jin, Yaohui and Lan, Zhenzhong and Kong, Lingpeng and He, Junxian},
  journal={Advances in neural information processing systems},
  volume={37},
  pages={74325--74362},
  year={2024}
}

@article{bo2024reflective,
  title={Reflective multi-agent collaboration based on large language models},
  author={Bo, Xiaohe and Zhang, Zeyu and Dai, Quanyu and Feng, Xueyang and Wang, Lei and Li, Rui and Chen, Xu and Wen, Ji-Rong},
  journal={Advances in Neural Information Processing Systems},
  volume={37},
  pages={138595--138631},
  year={2024}
}

@article{xu2025towards,
  title={Towards large reasoning models: A survey of reinforced reasoning with large language models},
  author={Xu, Fengli and Hao, Qianyue and Zong, Zefang and Wang, Jingwei and Zhang, Yunke and Wang, Jingyi and Lan, Xiaochong and Gong, Jiahui and Ouyang, Tianjian and Meng, Fanjin and others},
  journal={arXiv preprint arXiv:2501.09686},
  year={2025}
}

@inproceedings{qin2025meta,
  title={Meta-Tool: Unleash Open-World Function Calling Capabilities of General-Purpose Large Language Models},
  author={Qin, Shengqian and Zhu, Yakun and Mu, Linjie and Zhang, Shaoting and Zhang, Xiaofan},
  booktitle={Proceedings of the 63rd Annual Meeting of the Association for Computational Linguistics (Volume 1: Long Papers)},
  pages={30653--30677},
  year={2025}
}

@article{jimenez2023swe,
  title={Swe-bench: Can language models resolve real-world github issues?},
  author={Jimenez, Carlos E and Yang, John and Wettig, Alexander and Yao, Shunyu and Pei, Kexin and Press, Ofir and Narasimhan, Karthik},
  journal={arXiv preprint arXiv:2310.06770},
  year={2023}
}

@article{patil2024gorilla,
  title={Gorilla: Large language model connected with massive apis},
  author={Patil, Shishir G and Zhang, Tianjun and Wang, Xin and Gonzalez, Joseph E},
  journal={Advances in Neural Information Processing Systems},
  volume={37},
  pages={126544--126565},
  year={2024}
}

@article{shen2023hugginggpt,
  title={Hugginggpt: Solving ai tasks with chatgpt and its friends in hugging face},
  author={Shen, Yongliang and Song, Kaitao and Tan, Xu and Li, Dongsheng and Lu, Weiming and Zhuang, Yueting},
  journal={Advances in Neural Information Processing Systems},
  volume={36},
  pages={38154--38180},
  year={2023}
}

@article{jha2025itbench,
  title={Itbench: Evaluating ai agents across diverse real-world it automation tasks},
  author={Jha, Saurabh and Arora, Rohan and Watanabe, Yuji and Yanagawa, Takumi and Chen, Yinfang and Clark, Jackson and Bhavya, Bhavya and Verma, Mudit and Kumar, Harshit and Kitahara, Hirokuni and others},
  journal={arXiv preprint arXiv:2502.05352},
  year={2025}
}

@inproceedings{yao2023react,
  title={React: Synergizing reasoning and acting in language models},
  author={Yao, Shunyu and Zhao, Jeffrey and Yu, Dian and Du, Nan and Shafran, Izhak and Narasimhan, Karthik and Cao, Yuan},
  booktitle={International Conference on Learning Representations (ICLR)},
  year={2023}
}

@inproceedings{ye2024efficient,
  title={An Efficient Open World Benchmark for Multi-Agent Reinforcement Learning},
  author={Ye, Eric and Jaques, Natasha},
  booktitle={NeurIPS 2024 Workshop on Open-World Agents}
}

@article{sapkota2025ai,
  title={Ai agents vs. agentic ai: A conceptual taxonomy, applications and challenges},
  author={Sapkota, Ranjan and Roumeliotis, Konstantinos I and Karkee, Manoj},
  journal={arXiv preprint arXiv:2505.10468},
  year={2025}
}

@article{shinn2023reflexion,
  title={Reflexion: Language agents with verbal reinforcement learning},
  author={Shinn, Noah and Cassano, Federico and Gopinath, Ashwin and Narasimhan, Karthik and Yao, Shunyu},
  journal={Advances in Neural Information Processing Systems},
  volume={36},
  pages={8634--8652},
  year={2023}
}

@article{renze2024self,
  title={Self-reflection in llm agents: Effects on problem-solving performance},
  author={Renze, Matthew and Guven, Erhan},
  journal={arXiv preprint arXiv:2405.06682},
  year={2024}
}

@article{kapoor2024ai,
  title={Ai agents that matter},
  author={Kapoor, Sayash and Stroebl, Benedikt and Siegel, Zachary S and Nadgir, Nitya and Narayanan, Arvind},
  journal={arXiv preprint arXiv:2407.01502},
  year={2024}
}

@misc{masterman_tool_calling_2024,
    title = {AI Agents: The Intersection of Tool Calling and Reasoning in Generative AI},
    author = {Masterman, Tula},
    url = {https://medium.com/data-science/ai-agents-the-intersection-of-tool-calling-and-reasoning-in-generative-ai-ff268eece443},
    year = {2024},
    month = {10},
    urldate = {2025-08-20}
}

@article{plaat_artist_2025,
    title = {Agentic Reasoning and Tool Integration for LLMs via Reinforcement Learning},
    author = {Plaat, Aske and others},
    journal = {arXiv},
    year = {2025},
    month = {4},
    url = {https://arxiv.org/html/2505.01441v1},
    urldate = {2025-08-20}
}

@article{wu_large_reasoning_models_2025,
    title = {Towards Large Reasoning Models: A Survey of Reinforced Reasoning with Large Language Models},
    author = {Wu, Q. and others},
    journal = {arXiv},
    year = {2025},
    month = {1},
    url = {https://arxiv.org/html/2501.09686v3},
    urldate = {2025-08-20}
}

@misc{agentic_ai_frameworks_2025,
    title = {Agentic AI Frameworks: Architectures, Protocols, and Design Challenges},
    author = {{LlamaIndex}},
    url = {https://arxiv.org/html/2508.10146},
    year = {2025},
    month = {8},
    urldate = {2025-08-20}
}

@article{patent_ai_workflows_2024,
    title = {Towards Automated Patent Workflows: AI-Orchestrated Multi-Agent Framework for Intellectual Property Management and Analysis},
    author = {Zhang, Y. and others},
    journal = {arXiv},
    year = {2024},
    month = {10},
    url = {https://arxiv.org/html/2409.19006},
    urldate = {2025-08-20}
}

@misc{afp_openai_operator_2025,
    title = {OpenAI lance un agent IA autonome qui exécute des tâches en ligne},
    author = {{AFP}},
    url = {https://apnews.com/article/nvidia-gtc-jensen-huang-ai-457e9260aa2a34c1bbcc07c98b7a0555},
    year = {2025},
    month = {1},
    urldate = {2025-08-20}
}
